\documentclass[sigconf]{acmart}

\AtBeginDocument{%
  }

\copyrightyear{2024}
\acmYear{2024}
\setcopyright{rightsretained}
\acmConference[ICAIF '24]{5th ACM International Conference on AI in Finance}{November 14--17, 2024}{Brooklyn, NY, USA}
\acmBooktitle{5th ACM International Conference on AI in Finance (ICAIF '24), November 14--17, 2024, Brooklyn, NY, USA}
\acmDOI{10.1145/3677052.3698635}
\acmISBN{979-8-4007-1081-0/24/11}

\usepackage{url}
\usepackage{algpseudocode}
\usepackage{algorithm}
\usepackage{graphicx}
\usepackage{tabularx}
\usepackage{multirow}
\usepackage{subcaption}
\usepackage{graphicx}
\usepackage{booktabs} 
\usepackage{float} 
\usepackage{todonotes}
\usepackage{pgfplots}
\pgfplotsset{compat=1.17}

\newcolumntype{P}[1]{>{\centering\arraybackslash}p{#1}}

\usepackage{xspace}
\usepackage{enumitem}

\newcommand{\ellipticI}{\emph{Elliptic1}\xspace}
\newcommand{\ellipticII}{\emph{Elliptic2}\xspace}
\newcommand{\revtrack}{\emph{RevTrack}\xspace}
\newcommand{\revclassify}{\emph{RevClassify}\xspace}
\newcommand{\revclassifyBP}{\emph{RevClassify}$_{BP}$\xspace}
\newcommand{\revclassifyDS}{\emph{RevClassify}$_{DS}$\xspace}
\newcommand{\revfilter}{\emph{RevFilter}\xspace}

\newcommand{\vh}{\mathbf{h}}

\newcommand{\tC}{\mathcal{C}}
\newcommand{\tD}{\mathcal{D}}

\newcommand{\tG}{\mathcal{G}}
\newcommand{\tH}{\mathcal{H}}

\newcommand{\tS}{\mathcal{S}}

\newcommand{\sE}{\mathbb{E}}

\newcommand{\sR}{\mathbb{R}}
\newcommand{\sS}{\mathbb{S}}

\newcommand{\sV}{\mathbb{V}}

\begin{document}

\title{Identifying Money Laundering Subgraphs on the Blockchain}

\author{Kiwhan Song}
\authornote{Both authors contributed equally to this research.}
\email{kiwhan@mit.edu}
\orcid{0009-0000-7103-0301}
\affiliation{%
  \institution{MIT, MIT-IBM Watson AI Lab}
  \city{Cambridge}
  \state{MA}
  \country{USA}
}
\author{Mohamed Ali Dhraief}
\authornotemark[1]
\email{mohamed.ali.dhraief@ibm.com}
\orcid{0009-0003-0432-853X}
\affiliation{%
  \institution{IBM}
  \city{Zurich}
  \country{Switzerland}
}
\author{Muhua Xu}
\email{muhuaxu@mit.edu}
\orcid{0009-0007-5416-3533}
\affiliation{%
  \institution{MIT, MIT-IBM Watson AI Lab}
  \city{Cambridge}
  \state{MA}
  \country{USA}
}
\author{Locke Cai}
\email{lcai12@mit.edu}
\orcid{0009-0007-6440-5989}
\affiliation{%
  \institution{MIT, MIT-IBM Watson AI Lab}
  \city{Cambridge}
  \state{MA}
  \country{USA}
}
\author{Xuhao Chen}
\email{cxh@mit.edu}
\orcid{0000-0001-6470-3387}
\affiliation{%
  \institution{MIT, MIT-IBM Watson AI Lab}
  \city{Cambridge}
  \state{MA}
  \country{USA}
}
\author{Arvind}
\email{arvind@csail.mit.edu}
\orcid{0000-0002-9737-2366}
\affiliation{%
  \institution{MIT, MIT-IBM Watson AI Lab}
  \city{Cambridge}
  \state{MA}
  \country{USA}
}
\author{Jie Chen}
\email{chenjie@us.ibm.com}
\orcid{0000-0002-0449-6803}
\affiliation{%
  \institution{IBM Research, MIT-IBM Watson AI Lab}
  \city{Cambridge}
  \state{MA}
  \country{USA}
}

\renewcommand{\shortauthors}{Song et al.}

\begin{abstract}
  Anti-Money Laundering (AML) involves the identification of money laundering crimes in financial activities, such as cryptocurrency transactions. Recent studies advanced AML through the lens of graph-based machine learning, modeling the web of financial transactions as a graph and developing graph methods to identify suspicious activities. 
  For instance, a recent effort on opensourcing datasets and benchmarks, 
  \emph{Elliptic2}, treats a set of Bitcoin addresses, considered to be controlled by the same entity, as a graph node and transactions among entities as graph edges. This modeling reveals the ``shape'' of a money laundering scheme---a subgraph on the blockchain, such as a peeling chain or a nested service. 
  Despite the attractive subgraph classification results benchmarked by the paper, competitive methods remain expensive to apply due to the massive size of the graph; moreover, existing methods require candidate subgraphs as inputs which may not be available in practice.

  In this work, we introduce \emph{RevTrack}, a graph-based framework that enables large-scale AML analysis with a lower cost and a higher accuracy. The key idea is to track the initial senders and the final receivers of funds; these entities offer a strong indication of the nature (licit vs. suspicious) of their respective subgraph. Based on this framework, we propose \emph{RevClassify}, which is a neural network model for subgraph classification. Additionally, we address the practical problem where subgraph candidates are not given, by proposing \emph{RevFilter}. This method identifies new suspicious subgraphs by iteratively filtering licit transactions, using \emph{RevClassify}. Benchmarking these methods on \emph{Elliptic2}, a new standard for AML, we show that \emph{RevClassify} outperforms state-of-the-art subgraph classification techniques in both cost and accuracy. Furthermore, we demonstrate the effectiveness of \emph{RevFilter} in discovering new suspicious subgraphs, confirming its utility for practical AML.
  
\end{abstract}



\maketitle


\begin{figure*}[t]
  \centering
  \includegraphics[width=.95\linewidth]{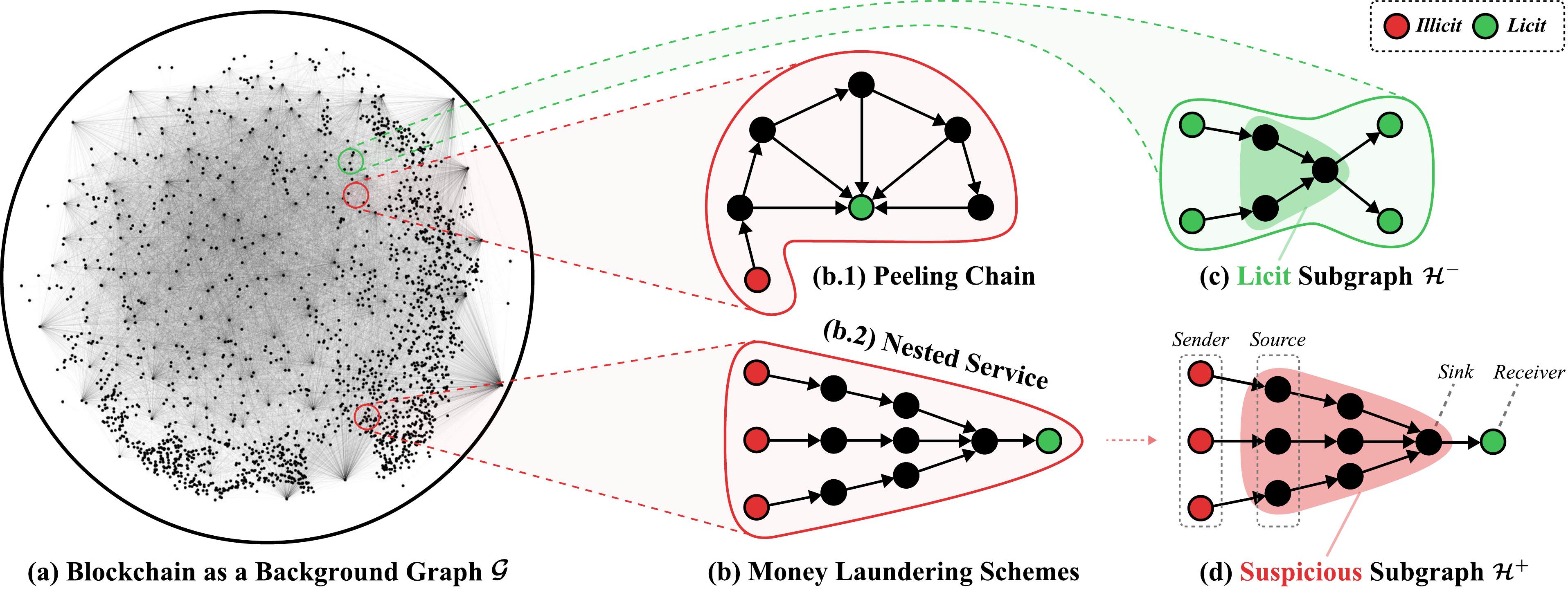}
  \caption{Terminology and examples of money laundering schemes.}
  \label{fig:subgraph.AML}
\end{figure*}

\section{Introduction}
Money laundering, a financial crime, supports a range of destructive activities including human trafficking, drug trafficking, and terrorism. Over 70\% of criminal networks use money laundering to fund their activities and conceal their assets~\cite{Europol2024}. This demonstrates not only the extensive reach of money laundering across various crimes but also the profound consequence, with estimated \$3.1 trillion in illicit funds flowed through the global financial system in 2023~\cite{Nasdaq2023}. The rise of cryptocurrencies has accelerated this issue, with the number of victims and volume of dollar lost to crypto-frauds doubling~\cite{FBI2022}. Worse, the pesudo-anonymous nature of the technology offers the illegal activities an additional layer of protection. Combating financial crimes today is crucial and is considered a matter of national security~\cite{Treasury2024}.

Driven by the need for better tools for anti-money laundering (AML), the Elliptic dataset (henceforth referred to as \ellipticI) was released in 2019, as the largest, publicly accessible, labeled AML/cryptocurrency dataset at the time~\cite{Weber2019}. \ellipticI is a graph, comprising 204K crypto transactions as graph nodes and 234K payment flows (the flows of every Bitcoin from one transaction to the next) as directed edges. The machine learning task is node classification---classifying the licit vs. illicit nature of each transaction. While \ellipticI gained significant traction in both the machine learning and the AML communities, the task it supports is limited to analyzing individual transactions, leaving the question of studying more complex money laundering schemes unaddressed.

Money laundering schemes involve a series of financial transactions that transform illegally obtained funds to apparently legal accounts. In 2024, \ellipticII was released in response to the need for effective tools to unveil such complex patterns on the cryptocurrency blockchain~\cite{Bellei2024}. Besides being a graph that is nearly three orders of magnitude larger than \ellipticI, \ellipticII models the transaction graph differently and features a different machine learning task---subgraph classification. Specifically, each node represents a financial entity on the blockchain and each edge aggregates transactions between a pair of entities. Thus, a money laundering scheme signifies a subgraph consisting of a series of transactions among multiple entities, likely from criminal to legal.

Representing money laundering schemes as subgraphs allows one to leverage the rapid progress of the field of graph-based machine learning~\cite{Zhou2020,Wu2021} to develop effective AML tools. In particular, subgraph classification is an emerging topic, which extends the thoroughly studied tasks---node classification, edge prediction, and graph classification---to yet another granularity of graphs---subgraphs. Several subgraph neural network methods~\cite{adhikari2018sub2vec,alsentzer2020subgraph,wang2021glass} appear to be good candidates.

However, there are multiple challenges in subgraph AML. First, crypto transaction graphs are massive. The cumulated number of transactions on the blockchain exceeded one billion on May 05, 2024~\cite{blockchain.com}. Even with address and transaction aggregation, \ellipticII includes nearly 50M nodes and 200M edges. 
Such a large scale costs effective subgraph models, such as GLASS~\cite{wang2021glass}, a few days to train without GPUs, while requiring nonstraightforward system engineering efforts to port the training on GPUs~\cite{Bellei2024}. 
Second, there are exponentially many subgraphs in a graph; identifying suspicious subgraphs corresponding to money laundering is like finding needles in a haystack. A typical classification method can classify a reasonable amount of instances, but it becomes impractical when the amount is exponential.

In this work, we develop a framework, coined \revtrack, which allows efficient classification and discovery of suspicious money laundering subgraphs. A key to this framework is to track the sending and receiving entities of the subgraph rather than the subgraph itself. Doing so allows the use of alternative neural networks (other than graph neural networks) that are easier to train and scale better. Based on this framework, we propose a \revclassify method for subgraph classification (classifying a given subgraph) and a \revfilter method for identifying potential criminal entities and their money laundering activities (discovering new, suspicious subgraphs). Benchmarking these methods on the \ellipticII dataset, we demonstrate their superior performance over strong baselines and their practical utility in AML.

\section{Subgraph Representations of Money Laundering}
A majority of the participants on the blockchain are ``licit'' entities; they range from exchanges, wallet providers, miners, to licit services. On the other hand, an ``illicit'' entity is commonly associated with crimes, such as dark markets, scammers, and hackers. A fundamental assumption of money laundering is that a path connecting an illicit entity to a licit one without a change of ownership of the funds likely represents money laundering by a criminal person or organization. Through layers of transactions (laundering), criminals deposit funds at a legitimate service and evade detection of the illegal source of the funds.

A money laundering scheme may consist of one or multiple such illict$\to$licit paths; the union of these paths is a subgraph. A known scheme is a ``peeling chain'' (see the middle illustration of Figure~\ref{fig:subgraph.AML}), where all the intermediate entities on the path additionally point to the end of the path. In this case, the ending entity could be an exchange, to which all intermediate entities deposit part of the funds (with the rest sent to the next entity). Another example is a ``nested service'' (see also Figure~\ref{fig:subgraph.AML}), where multiple paths starting from separate illicit entities merge on the same ``service'' entity, which further points to an exchange that is licit. Such services typically have less stringent customer due diligence checks than the exchanges, resulting in their (mis)use for cryptocurrency laundering.

\subsection{Terminology and Notation}\label{sec:def}
We illustrate in Figure~\ref{fig:subgraph.AML} the main concepts used throughout the paper.
The blockchain is modeled as a directed graph $\tG = (\sV, \sE)$, where a node $v \in \sV$ is a set of Bitcoin addresses thought to be controlled by a single entity (e.g., a person or organization) and a directed edge $e = (u, v) \in \sE$ denotes one or multiple transactions from entity $u$ to $v$. We call $\tG$ the \emph{background graph} when its subgraphs are of concern.

Denote by $\tH = (\sV_{\tH}, \sE_{\tH})$ a subgraph of $\tG$. Each node $v \in \sV_{\tH}$ with a zero in-degree inside $\tH$ is called a \emph{source} and all sources form the set $\sV_{source}$. Similarly, each node with a zero out-degree inside $\tH$ is called a \emph{sink} and all sinks form the set $\sV_{sink}$. A node pointing to any source is called a \emph{sender} and all senders form the set $\sS$. Similarly, a node that is pointed to by any sink is called a \emph{receiver} and all receivers form the set $\sR$. Note that by definition, $\sS$ and $\sR$ are outside the subgraph $\tH$.

The \ellipticII dataset constructs subgraphs for classification with the help of node labeling that is undisclosed. A node is (manually) labeled as \emph{licit}, \emph{illicit}, or in most of the cases, unlabeled, in which case we call it \emph{unknown}. Subgraphs in \ellipticII are labeled either \emph{licit} or \emph{suspicious}. Based on the construction procedure described in~\cite{Bellei2024}, one can infer that a subgraph in \ellipticII is licit if the senders and receivers are all licit, while a subgraph is suspicious if the receivers are licit but the senders are illicit. Suspicious subgraphs are intended to be validated by human analysts to confirm their illicit nature (money laundering).

Note that not every subgraph can be labeled; and illicit subgraphs are scarce among those labeled. It is tempting to predict the subgraph labels based on the labels of the senders and receivers. However, node labels are not provided and reverse engineering them based on the subgraph construction procedure will reveal the labels of only a small fraction of the nodes.

\subsection{Two Tasks of Interest}
In this work, we are interested in two tasks:
\begin{enumerate}[leftmargin=*]
\item Classify the nature (licit vs. suspicious) of a given subgraph $\tH$;
\item Identify new suspicious subgraphs.
\end{enumerate}

Task (1) is a standard classification problem: Given a set of subgraphs with labels, split it into training/validation/test subsets. Train a model to predict the label of each subgraph by using the training and validation sets and evaluate it on the test set. This task solicits effective machine learning models that can discern the nature of any given subgraph.

The ultimate goal of AML, on the other hand, is to discover all money laundering schemes. Task (1) is infeasible for this goal, for two reasons. First, there are exponentially many (specifically, $2^{|\sV|}$) subgraphs; it is impossible to enumerate them and classify one by one. Second, not every subgraph can be logically labeled; in fact, only a few can be considered licit and even few are suspicious.

Hence, Task (2) is an alternative solution to this goal: Given a set of labeled subgraphs, design a method that can discover new subgraphs that are likely suspicious (among the exponentially many candidates). This task is highly nontrivial. We propose a method by reusing the learned model from Task (1) to accomplish the task.

\section{Identifying Money Laundering Subgraphs with Senders \& Receivers}

\textbf{Subgraph neural networks are costly.} A straightforward approach to telling if a subgraph is a suspicious money laundering scheme is to perform subgraph classification. Several representative neural networks for this approach were explored in~\cite{Bellei2024}, suggesting that the effective ones for our use case are quite expensive to train. For example, GLASS~\cite{wang2021glass}, using as few as two layers, requires several days to train by using CPUs. 
This is because GPU training is a complex matter for a background graph as large as \ellipticII ($\sim$50M nodes and $\sim$200M edges) due to the massive memory consumption
and it requires nonstraightforward engineering efforts to adapt the node classification workload~\cite{kaler2022accelerating,kaler2023communication} to subgraph classification workload~\cite{Bellei2024}.
While GPU training is the standard for usual neural networks, for graph neural networks it requires special handling because the loss of a data point (i.e., a node) requires the information of not just the data itself but also its neighborhood, which causes debates regarding whether full-graph training or mini-batch training should be used. Moreover, if one uses mini-batch training, adapting batching and neighborhood sampling to subgraphs requires reengineering existing libraries or codebases for effective memory usage~\cite{kaler2022accelerating,kaler2023communication}.

\begin{figure}[t]
  \centering
  \includegraphics[width=.9\linewidth]{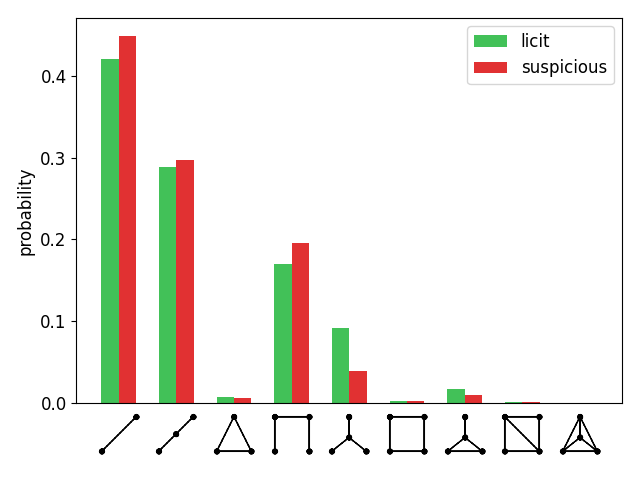}
  \caption{Graphlet distribution for licit subgraphs and that for suspicious subgraphs.}
  \label{fig:graphlet}
\end{figure}

\textbf{The subgraph structure alone is insufficient for identification.} A reason why GLASS, a method that trains a neural network on the entire background graph, performs more effective classification than do methods that train a neural network on individual subgraphs only (e.g., Sub2Vec~\cite{adhikari2018sub2vec}), is that the internal subgraph structure alone is insufficient for classification. What matters additionally is the border information surrounding but outside the subgraph. To further illustrate this point, we compute the distribution of graphlets (including 2-node, 3-node, and 4-node graphlets) inside a collection of subgraphs. Figure~\ref{fig:graphlet} shows that the distribution for licit subgraphs and that for suspicious ones are rather similar. Although this example is simplified for computational purpose (e.g., ignoring edge directions and larger graphlets), more evidences from the subsequent experiment section confirm that one should look beyond the internal structures for identifying suspicious subgraphs.

\textbf{Senders and receivers of funds provide a strong hint.} A logically useful piece of border information is the senders and receivers, because field wisdom suggests that illicit senders launder money through transferring funds layer by layer to licit accounts. Hence, in the next section, we propose methods to identify money laundering subgraphs with a focal use of sender and receiver information. These entities can be easily extracted by using the definition introduced in Section~\ref{sec:def} (see also the example in Figure~\ref{fig:subgraph.AML}): For each subgraph, we first identify nodes where transactions begin and end, namely, \emph{sources} and \emph{sinks}. Then, the nodes outside the subgraph pointing to the sources are \emph{senders} and those being pointed to by the sinks are \emph{receivers}. Note that a subgraph occasionally is not acyclic (e.g., two entities may send funds to each other at different times), which leads to no sources or sinks. In this case, we remove all detected cycles to extract sources and sinks.

\begin{figure*}[t]
  \centering
  \includegraphics[width=1.0\linewidth]{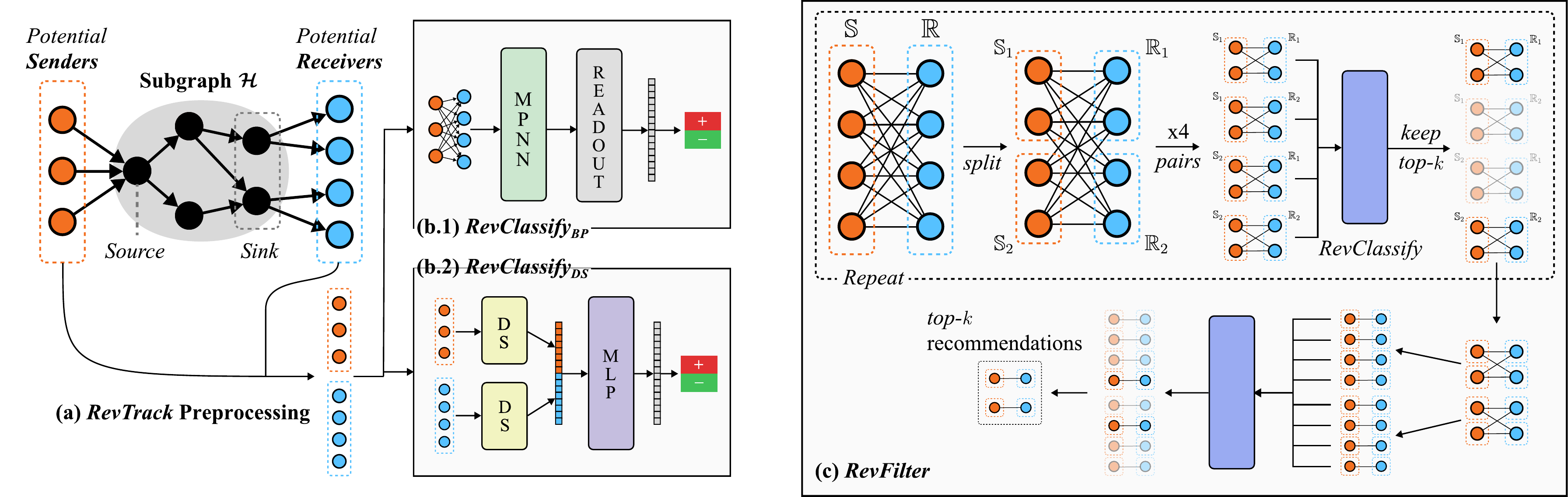}
  \caption{Illustration of the \revclassify and \revfilter methods.}
  \label{fig:RevTrack}
\end{figure*}

\section{\revtrack: A Tale of Two Methods}
We are now ready to introduce \revtrack, a framework that analyzes and discovers money laundering subgraphs through tracking senders and receivers. The framework represents a subgraph $\tH$ by its sender set $\sS$ and its receiver set $\sR$. We create ``links'' between them; these links can be interpreted as an abstraction of the paths between $\sS$ and $\sR$. \revtrack includes two methods: \revclassify classifies the nature (licit vs. suspicious) of a given subgraph, while \revfilter discovers new suspicious subgraphs through iteratively filtering out deemed licit links. Figure~\ref{fig:RevTrack} illustrates both methods.

\subsection{\revclassify: Subgraph Classification}
\revclassify classifies a given subgraph $\tH$. Given the $(\sS,\sR)$ representation of $\tH$, the method generates an embedding vector $\vh_{\tH}$ for the subgraph, such that a logistic regression of $\vh_{\tH}$ performs the classification. \revclassify requires an expressive architecture to handle the two sets as inputs. Here, we name two architectures, \revclassifyBP and \revclassifyDS, both of which are permutation-invariant with respect to $\sS$ and $\sR$.

\subsubsection*{\revclassifyBP}
This architecture builds a fully-connected, directed bipartite graph $\widetilde{\tH}$ between $\sS$ and $\sR$. That is,
\begin{equation}
\widetilde{\tH} = (\widetilde{\sV}_{\tH}, \widetilde{\sE}_{\tH}),
\end{equation}
where $\widetilde{\sV}_{\tH} = \sS \cup \sR$ and $\widetilde{\sE}_{\tH} = \{ (u,v) \mid u \in \sS \land v \in \sR \}$. Then, the embedding $\vh_{\tH}$ is computed by using any Message Passing Neural Network (MPNN)~\cite{kipf2016semi, xu2018powerful, velivckovic2017graph}, followed by a global pooling layer as the readout:
\begin{equation}
\vh_{\tH} = \text{READOUT} \left( \text{MPNN}(\widetilde{\tH}) \right).
\end{equation}
This architecture effectively runs a graph neural network on an alternative but augmented version of the subgraph, $\widetilde{\tH}$, to classify $\tH$. It considers information outside $\tH$ and is more effective.

\subsubsection*{\revclassifyDS}
This architecture does not use any graph neural network. Instead, it processes $\sS$ and $\sR$ separately by using Deep Sets~\cite{zaheer2017deep}, a universal architecture that generates set representations preserving permutation invariance and equivariance:
\begin{equation}
\vh_{\sS} = \text{DeepSets}(\sS) \quad \vh_{\sR} = \text{DeepSets}(\sR).
\end{equation}
This is suitable for our setting as there is no preference among the senders or receivers. Then, $\vh_{\sS}$ and $\vh_{\sR}$ are concatenated and a Multi-Layer Perceptron (MLP) is used to yield the final embedding:
\begin{equation}
\vh_{\tH} = \text{MLP} \left( \text{CONCAT}(\vh_{\sS}, \vh_{\sR}) \right).
\end{equation}
Both \revclassifyBP and \revclassifyDS can learn to classify $\tH$ effectively when trained with the binary cross-entropy loss.

\subsection{\revfilter: Discovering New Suspicious Subgraphs}\label{sec:discover}

\begin{algorithm*}[t]
\caption{\revfilter: Discovering New Suspicious Subgraphs}
\label{alg:rec}
\begin{algorithmic}[1]
    \State \textbf{Input:} A pair of sender set and receiver set, $(\mathbb{S}, \mathbb{R})$; Number of pairs to recommend, $k$; Pretrained classifier $\mathcal{C}$
    \State \textbf{Output:} Top-$k$ ordered list of suspicious sender-set-and-receiver-set pairs, $L$
    \State $L \gets \left[(\mathbb{S}, \mathbb{R})\right]$
    \Repeat
        \For{$(\mathbb{S}_i, \mathbb{R}_i) \in L$}
            \State $\mathbb{S}_i \rightarrow \mathbb{S}_i^1 + \mathbb{S}_i^2$, $\mathbb{R}_i \rightarrow \mathbb{R}_i^1 + \mathbb{R}_i^2$ \Comment{split senders/receivers into two equal-sized groups}
            \State $L \gets L - \left[(\mathbb{S}_i, \mathbb{R}_i)\right] + \left[(\mathbb{S}_i^1, \mathbb{R}_i^1), (\mathbb{S}_i^1, \mathbb{R}_i^2), (\mathbb{S}_i^2, \mathbb{R}_i^1), (\mathbb{S}_i^2, \mathbb{R}_i^2)\right] $ \Comment{replace the pair $(\mathbb{S}_i, \mathbb{R}_i)$ with four new pairs}
        \EndFor
        \If{$\lvert L \rvert > k$}
            \State scores $s \gets \mathcal{C}(L)$ then $\text{REVERSE-SORT}(L, \text{key}=s)$ 
            \Comment{sort the list $L$ by the scores computed by the classifier $\mathcal{C}$}
            \State $L \gets L[:k]$ \Comment{keep only the top-$k$ scoring pairs}
        \EndIf
        \Until{$\lvert L \rvert = k$ and $\forall(\mathbb{S}_i, \mathbb{R}_i) \in L, \lvert\mathbb{S}_i\rvert = \lvert\mathbb{R}_i\rvert = 1$ } \Comment{repeat until all $k$ pairs are 1-1}
        \State{\textbf{return} $L$}
\end{algorithmic}
\end{algorithm*}

Although \revclassify can accurately classify subgraphs, it assumes that the subgraphs of interest are provided. However, there are in total $2^{|\sV|}$ subgraphs and only a tiny fraction of them correspond to money laundering. Hence, it is impractical to enumerate all subgraphs and apply \revclassify on them one by one. Rather, there is a strong desire for a recommendation-like system that can efficiently discover potential money laundering without exhaustive search.

To this end, we propose \revfilter, an efficient and scalable method that recommends suspicious subgraphs. The method works by iteratively filtering out groups of senders and receivers that are deemed to not contribute to money laundering. Specifically, given an arbitrary set of senders, $\sS$, and an arbitrary set of receivers, $\sR$, \revfilter produces a list of top-$k$ ranked $(s,r)$ pairs, where $s \in \sS$, $r \in \sR$, and each $(s,r)$ pair is deemed suspicious (i.e., the path connecting $s$ and $r$ is a suspicious subgraph). Remarkably, \revfilter leverages a pretrained \revclassify classifier, thus eliminating the need for any training or optimization.

Algorithm~\ref{alg:rec} describes the details. \revfilter maintains a list $L$ of sender-set-and-receiver-set pairs. Initially, there is only one pair, $(\sS,\sR)$. In each iteration, we split each pair in the list into four pairs, by bisecting the sender set and the receiver set. Then, all pairs are passed into the pretrained classifier $\tC$, which assigns a probability score indicating the likelihood of money laundering between the senders and the receivers. Based on the score, only the top-$k$ ranked pairs are retained for the next iteration. This iterative filtering process is repeated until the size of each pair is reduced to 1-1, suggesting top-$k$ most likely money laundering paths.

This basic version of \revfilter can be made more robust with the following enhancements, when suspicious subgraphs are scarce.

\textbf{(1) Fine-tuning with data augmentation.}
Note that the pairs passed into the pretrained classifier $\tC$ have a large sender/receiver set at the beginning. However, this classifier is mostly trained with small pairs, due to the nature of the \ellipticII dataset. Thus, to enhance the performance of $\tC$, we fine-tune it by using \emph{merged} pairs. These pairs are created by randomly merging sets of senders and receivers, such that the number of sets merged, $n_{merge}$, follows the exponential distribution
$P(n_{merge} = t) = \gamma e^{-\gamma \cdot t}$.
Such data augmentation generates a fine-tuning dataset where the size distribution of the pairs is similar to that encountered in the execution of \revfilter.

\textbf{(2) Keeping more candidates.}
If the classifier $\tC$ mistakenly assigns a low score to a pair that contains suspicious money laundering activities during iterations, the suspicious activities are erroneously ignored. Even when the classifier is highly accurate, this could occur through many iterations. Thus, for the initial iteration, we maintain $\alpha_{keep} \times k$ pairs ($\alpha_{keep} > 1$) and gradually decrease this number to $k$ by the end of the iterations, to mitigate accidentally eliminating suspicious pairs. Note that using a large $\alpha_{keep}$ may lead to a trade-off: while it helps preserve potential candidates, it can also result in a decrease in inference speed and a flattening of the iterative filtering process. 

\subsubsection*{Practical Use of RevFilter}
Algorithm~\ref{alg:rec} requires an initial sender set $\sS$ and a receiver set $\sR$ as input. 
One practical use of \revfilter is to identify additional illicit entities given known money laundering subgraphs. The end of these subgraphs are linked to licit entities (such as an exchange). Hence, using these entities to form $\sR$ potentially identify new money laundering schemes sharing the same choice of crypto deposits.
Additionally, one may partition the node set $\sV$ such that one partition is used to form $\sS$ each time, balancing the relative sizes of $\sS$ and $\sR$ while speeding up the inference.

\section{Experiments: \revclassify}
\label{sec:exp_classification}

We comprehensively evaluate the effectiveness of the proposed methods---\revclassify in this section and \revfilter in the next---with the \ellipticII dataset~\cite{Bellei2024}. \ellipticII was a recently released benchmark, the largest of its kind, which innovatively models AML as a subgraph classification problem. In the dataset, the background graph contains 49,299,864 nodes and 196,215,606 edges. Additionally, there are 121,810 labeled subgraphs, among which 119,092 are licit and 2,718 are suspicious.

\addtolength{\tabcolsep}{-0.33em}
\begin{table*}[t]
\centering
\caption{Accuracy and cost comparisons between \revclassify and baselines. ``Full-shot'' uses the whole training set for training while ``few-shot'' uses a portion of it. The best results are \textbf{boldfaced} and the second-best are \underline{underlined}.}
\label{tab:classification}
\begin{tabularx}{\textwidth}{X|cccccc|cc|ccc|c}
\toprule
\multirow{3.7}{*}{\textbf{Method}} & \multicolumn{6}{c|}{\textbf{Few-shot}}  & \multicolumn{2}{c|}{\textbf{Full-shot}} & \multicolumn{4}{c}{\textbf{Cost}} \\
\cmidrule{2-13}
{} & \multicolumn{2}{c}{3\%} & \multicolumn{2}{c}{10\%} & \multicolumn{2}{c|}{30\%} & \multicolumn{2}{c|}{} & \multicolumn{3}{c|}{Time (min)} & Memory (GB) \\
{} & PR-AUC $\uparrow$& F1 $\uparrow$& PR-AUC $\uparrow$& F1 $\uparrow$& PR-AUC $\uparrow$& F1 $\uparrow$& PR-AUC $\uparrow$ & F1 $\uparrow$ & Preprocess $\downarrow$& Train $\downarrow$& Inference $\downarrow$ & Train $\downarrow$ \\
\midrule
Sub2Vec & $0.024$ & $0.045$ & $0.024$ & $0.043$ & $0.024$ & $0.043$ & $0.025$ & $0.044$ & 105.4 & $\boldsymbol{2.1}$ & 0.037 & $2.00$\\
GNN-SEG & $0.098$ & $0.109$ & $0.170$ & $0.142$ & $0.194$ & $0.153$ & $0.413$ & $0.206$ & \textbf{7.5} & $75.4$ & 1.740 & $1.81$ \\
GNN-PLAIN & $0.158$ & $0.105$ & $0.327$ & $0.118$ & $0.440$ & $0.135$ & $0.789$ & $0.660$ & N/A & $25.7$ & 0.069 & $5.80$ \\
GLASS & $0.164$ & $0.107$ & $0.350$ & $0.134$ & $0.477$ & $0.149$ & $0.816$ & $0.705$ & N/A & $56.1$ & 0.070 & $6.23$ \\
\midrule
\revclassifyBP &
$\underline{0.235}$ & 
$\underline{0.140}$ & 
$\underline{0.372}$ & 
$\underline{0.157}$ & 
$\underline{0.599}$ & 
$\underline{0.319}$ & 
$\underline{0.972}$ & 
$\boldsymbol{0.954}$ &
\multirow{2}{*}{$\underline{40.4}$} &
$5.4$ & $\underline{0.006}$ & $\boldsymbol{0.49}$ \\
\revclassifyDS & $\boldsymbol{0.445}$ & $\boldsymbol{0.287}$ & $\boldsymbol{0.665}$ & $\boldsymbol{0.363}$ & $\boldsymbol{0.802}$ & $\boldsymbol{0.614}$ & $\boldsymbol{0.974}$ & $\underline{0.953}$ & {} & $\underline{3.5}$ & \textbf{0.005} & $\underline{0.51}$ \\
\bottomrule
\end{tabularx}
\end{table*}

\subsection{Experiment Setup}
Following~\cite{Bellei2024}, we split the labeled subgraphs into training, validation, and test sets randomly (80:10:10). In addition to the \textbf{full-shot} setting, which uses the entire training set for training, we also explore the \textbf{few-shot} setting, where only a fraction of the training set is used, to investigate model behaviors in data-scarce environments. For a given fraction $p$, we randomly sample $p$ of the suspicious subgraphs as well as licit subgraphs.

\subsubsection{Baselines.}
We compare \revclassify with four typical subgraph classification methods. \textbf{Sub2Vec}~\cite{adhikari2018sub2vec} is an early graph embedding method, which samples random walks within a subgraph and uses Paragraph2Vec to learn subgraph embeddings from the sampled walks.
\textbf{GNN-SEG} is an MPNN that also acts on only the internal structure of the subgraph. \textbf{GNN-PLAIN} is another MPNN, but the message passing is conducted on the background graph, such that the subgraph representation draws information additionally outside the subgraph. \textbf{GLASS}~\cite{wang2021glass} extends GNN-PLAIN by employing a zero-one labeling trick that distinguishes the nodes inside and outside the subgraph; this method is theoretically proved to be more expressive than GNN-PLAIN.

\subsubsection{Evaluation Metrics.}
Due to the label imbalance and the importance of the suspicious class, we use the binary F1-score (treating suspicious as positive) and the PR-AUC score to evaluate model performance.
Additionally, we compare the resource demands of each method, including time and memory usage.

\subsubsection{Implementation Details.}
\revclassifyBP employs GIN~\cite{xu2018powerful} as the MPNN backbone. \revclassifyDS uses MLP for the invariant layers in Deep Sets~\cite{zaheer2017deep}. GLASS
and Sub2Vec
are implemented from the codebases provided by the original authors. GNN-SEG and GNN-PLAIN use the same MPNN backbone as does GLASS.
We train all models to minimize the binary cross entropy loss using the Adam optimizer~\cite{kingma2014adam} for 1000 epochs for the baseline methods and 150 epochs for \revclassify, incorporating early stopping. We tune hyperparameters, including the number of layers, hidden dimensions, type of pooling, learning rate, dropout, and batch size.

The performance of the baselines Sub2Vec, GNN-SEG, and GLASS is (sometimes substantially) improved over that reported by~\cite{Bellei2024}. The results in~\cite{Bellei2024} were obtained by CPU training and ignoring node features, but our results are obtained by GPU training and leveraging node features, whenever possible. Additionally, for GLASS and GNN-PLAIN, we use only one MPNN layer with neighborhood sampling, which significantly reduces the training time and the memory requirement. Moreover, we implement data preprocessing to enable faster data loading and replace the memory-intensive GraphNorm layer~\cite{cai2021graphnorm} with the more light-weight LayerNorm~\cite{ba2016layer}. These modifications boost the speed and accuracy of the baselines, making them stronger competitors. Our experiments are conducted by using a single V100 GPU with 16GB VRAM.

\subsection{Results}
\label{sec:classification}

\subsubsection{Accuracy.}
The results are summarized in Table~\ref{tab:classification}. \revclassify outperforms all baselines across the board. Among the baselines, Sub2Vec performs the most poorly, because node features are not used. GNN-PLAIN and GLASS perform significantly better than GNN-SEG, confirming the importance of involving external node information beyond the internal subgraph structure. Our method further significantly improves over GNN-PLAIN and GLASS. Between the two architectures of our method, \revclassifyDS is robustly stronger in the few-shot settings.

\subsubsection{Cost.}
As shown by Table~\ref{tab:classification}, \revclassify is considerably more memory-efficient and much faster in inference than the other methods. This efficiency stems from the use of only sender/receiver nodes, whereas GNN-SEG computes with all nodes in the subgraph, and Sub2Vec, GLASS, and GNN-PLAIN involve the background graph. GNN-SEG, GNN-PLAIN, and GLASS are slow to train; on the other hand, Sub2Vec is fast, but it requires extensive preprocessing. The preprocessing of \revclassify is not as demanding as Sub2Vec. Moreover, processed senders and receivers are stored in a hash table for reuse and their costs are amortized over subgraphs.

\section{Experiments: \textit{RevFilter}}

In this section, we evaluate \revfilter for discovering new suspicious subgraphs. The objective is to answer the following questions.

\begin{itemize}[leftmargin=*]
\item \textbf{Q1}: How does \revfilter perform compared to baselines?  
\item \textbf{Q2a}: How does the sparsity of money laundering affect the performance of \revfilter?
\item \textbf{Q2b}: How many recommendations ($k$) are needed for \revfilter to discover the majority of money laundering activities?
\item \textbf{Q3}: (Ablation) Is the iterative filtering in Algorithm~\ref{alg:rec}, as opposed to a one-pass top-$k$ selection, necessary?
\item \textbf{Q4a}: (Ablation) How does fine-tuning with data augmentation impact the performance of \revfilter?
\item \textbf{Q4b}: (Ablation) How does the keeping of more candidates impact the performance of \revfilter?
\end{itemize}

\addtolength{\tabcolsep}{-0.04em}
\begin{table*}[t]
\centering
\caption{Performance comparison between \revfilter and baselines in eight settings ($n^+ + n^- @ k$). The two metrics HR and NDCG are the higher the better.}
\label{tab:recommendation}
\begin{tabularx}{\textwidth}{X|cccccccccccccccc}
\toprule
$\boldsymbol{n^++n^-@k}$
& \multicolumn{2}{c}{1+5@1}
& \multicolumn{2}{c}{1+10@1}
& \multicolumn{2}{c}{1+10@3}
& \multicolumn{2}{c}{1+100@3}
& \multicolumn{2}{c}{3+100@10}
& \multicolumn{2}{c}{3+1000@10}
& \multicolumn{2}{c}{10+1000@100}
& \multicolumn{2}{c}{10+10000@100} \\

Density (\%)
& \multicolumn{2}{c}{12.54}
& \multicolumn{2}{c}{6.83}
& \multicolumn{2}{c}{6.83}
& \multicolumn{2}{c}{0.92}
& \multicolumn{2}{c}{0.98}
& \multicolumn{2}{c}{0.16}
& \multicolumn{2}{c}{0.18}
& \multicolumn{2}{c}{0.02} \\

{}
& HR
& NDCG
& HR
& NDCG
& HR
& NDCG
& HR
& NDCG
& HR
& NDCG
& HR
& NDCG
& HR
& NDCG
& HR
& NDCG \\

\midrule
MLP
& 0.4935
& 0.4935
& 0.3151
& 0.3151
& $\underline{0.7383}$
& 0.5596
& 0.0156
& 0.0105
& 0.0499
& 0.0278
& 0.0000
& 0.0000
& 0.0088
& 0.0034
& 0.0000
& 0.0000 \\
NGCF
& $\underline{0.5560}$
& $\underline{0.5560}$
& $\underline{0.4831}$
& $\underline{0.4831}$
& 0.6393
& 0.5723
& $\underline{0.4570}$
& $\underline{0.4226}$
& $\underline{0.4199}$
& $\underline{0.3705}$
& $\underline{0.3516}$
& $\underline{0.3291}$
& $\underline{0.4008}$
& $\underline{0.3374}$
& $\underline{0.3590}$
& $\underline{0.3170}$ \\
LightGCN
& 0.5391
& 0.5391
& 0.4596
& 0.4596
& 0.6862
& $\underline{0.5925}$
& 0.4310
& 0.3661
& 0.3652
& 0.3139
& 0.2452
& 0.2150
& 0.2958
& 0.2556
& 0.1939
& 0.1853 \\
\midrule
\revfilter
& $\boldsymbol{0.9479}$
& $\boldsymbol{0.9479}$
& $\boldsymbol{0.9479}$
& $\boldsymbol{0.9479}$
& $\boldsymbol{0.9779}$
& $\boldsymbol{0.9424}$
& $\boldsymbol{0.9271}$
& $\boldsymbol{0.8208}$
& $\boldsymbol{0.8826}$
& $\boldsymbol{0.6298}$
& $\boldsymbol{0.7385}$
& $\boldsymbol{0.5291}$
& $\boldsymbol{0.8568}$
& $\boldsymbol{0.4569}$
& $\boldsymbol{0.6646}$
& $\boldsymbol{0.3790}$ \\
\bottomrule
\end{tabularx}
\end{table*}

\begin{figure*}[t]
  \centering
  \begin{subfigure}{0.498\textwidth}
    \centering
    \includegraphics[width=1.0\textwidth]{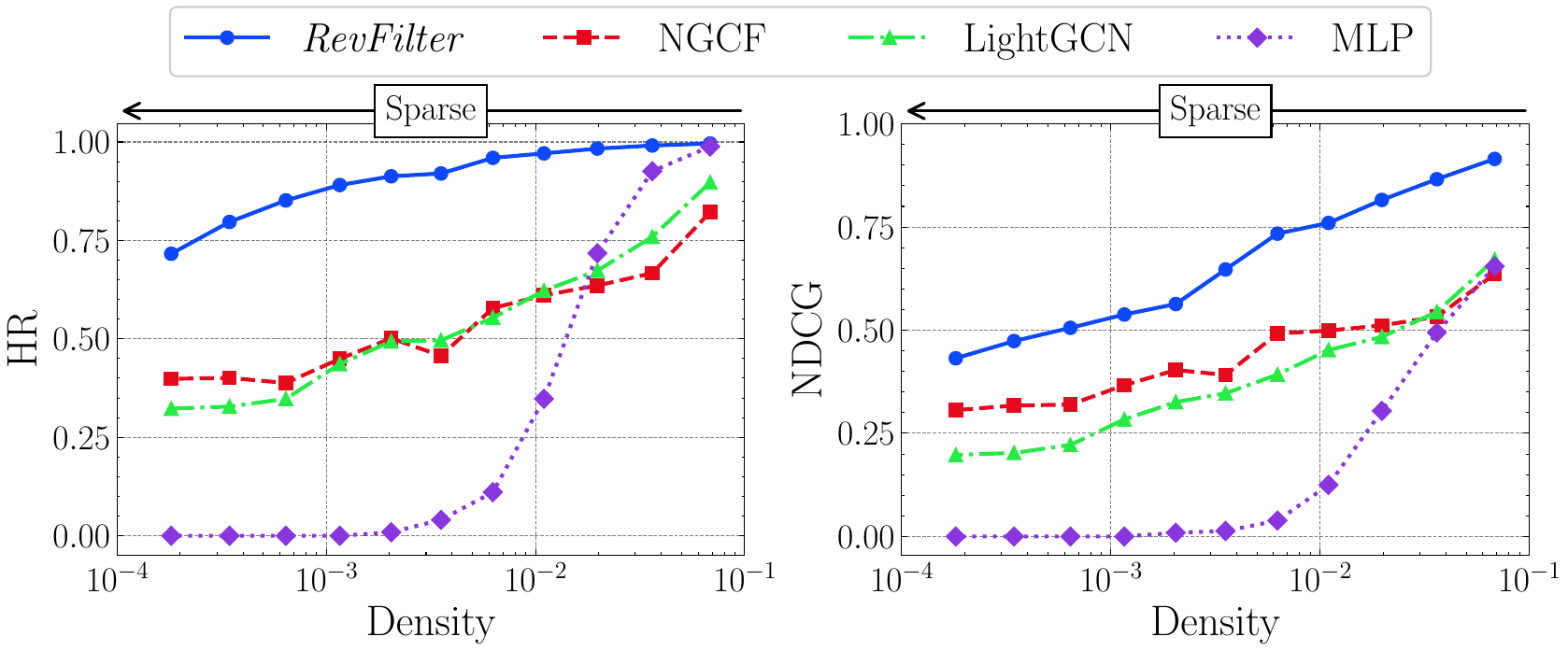}
    \caption{Impact of sparsity (in log-scale)}
    \label{fig:rec_plot_sparsity}
  \end{subfigure}
  \begin{subfigure}{0.498\textwidth}
    \centering
    \includegraphics[width=1.0\textwidth]{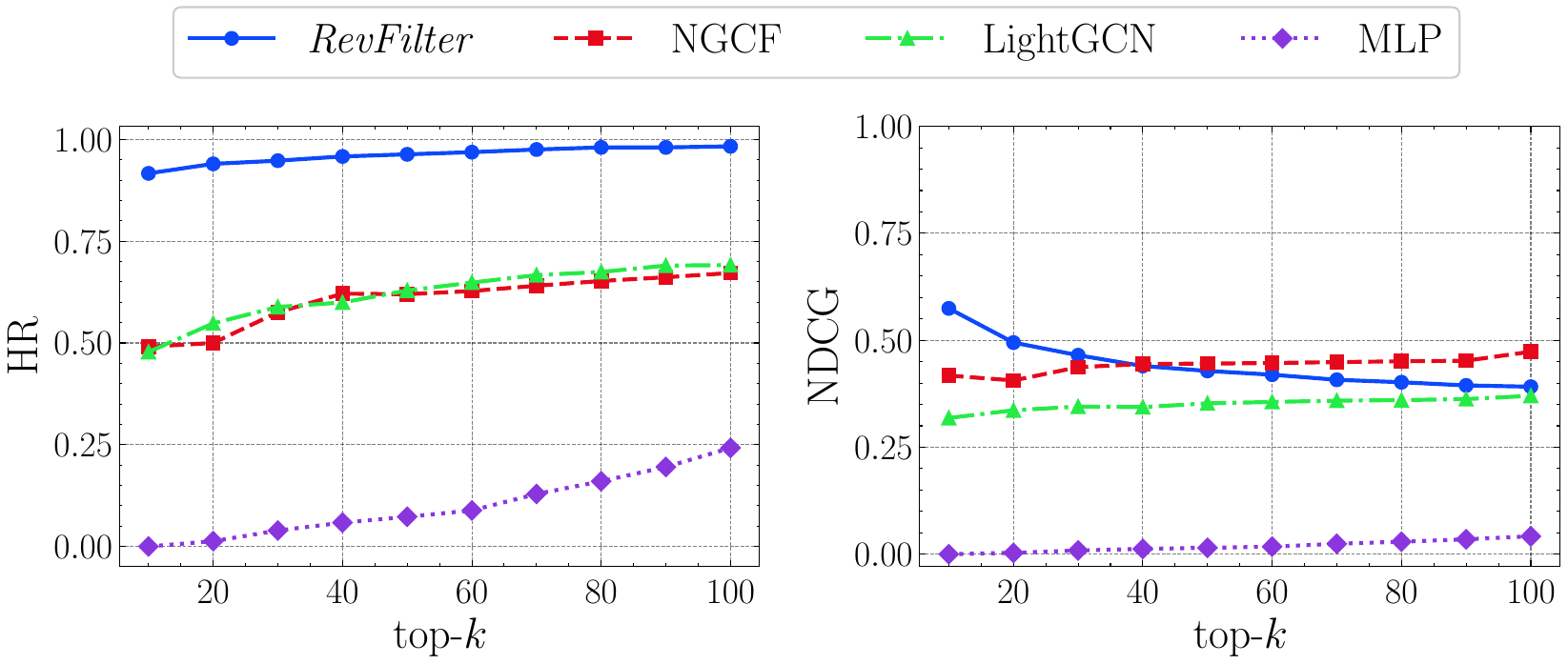}
    \caption{Impact of $k$}
    \label{fig:rec_plot_k}
  \end{subfigure}
  \caption{Comparison between \revfilter and baselines regarding sparsity and $k$.}
  \label{fig:rec_plot}
\end{figure*}

\subsection{Experiment Setup}
To the best of our knowledge, there does not exist previous literature experimenting on the same setting. The closest domain is collaborative filtering, which typically recommends items given a user (i.e., \textbf{single-user $\to$ multi-item}). In contrast, our task involves recommending links (suspicious 1-1 sender-receiver pairs) in a \textbf{multi-sender $\to$ multi-receiver} setting. Therefore, we adopt the experimental setup and evaluation protocol used in collaborative filtering and recommendation systems~\cite{he2017neural, wang2019neural, he2020lightgcn, ying2018graph} but make modifications to suit our task.

\subsubsection{Constructing a Test Set.}
We modify the test set of \ellipticII for the recommendation task, which recommends links between senders and receivers corresponding to suspicious money-laundering subgraphs. The modification proceeds as follows:

\begin{enumerate}[leftmargin=*]
\item For every subgraph, identify the senders and receivers. Denote the suspicious set as $\tD^+$ and the licit set as $\tD^-$.

\item Filter the suspicious set to include only subgraphs that induce a single link between senders and receivers (i.e., subgraphs with only a single sender and receiver). Call this subset $\tD^+_{1-1}$.

\item Randomly select $n^+$ subgraphs (links) from $\tD^+_{1-1}$ and $n^-$ subgraphs from $\tD^-$. Then, collect all the nodes from these subgraphs to construct a fully-connected, bipartite graph $\tS$ through linking the senders and receivers. The task is to recommend the $n^+$ links in $\tS$. \label{enum.line:s}

\item Repeat (\ref{enum.line:s}) $N$ times to construct a test set of size $N$. A large $N$ can significantly reduce the performance variance of a method. We take $N=256$ for all experiments.
\end{enumerate}

We evaluate each method by asking it to recommend top-$k$ links and denote the setting by $\boldsymbol{n^++n^-@k}$. The task is more challenging with a smaller $n^+$ and larger $n^-$, or with a smaller $k$. We define the \textbf{density} to be $n^+/(|\sS||\sR|)$ and call the case of small density, \textbf{sparse}. We evaluate on different settings with varying $n^+$, $n^-$, and $k$.

\subsubsection{Baselines.}
We compare \revfilter against three baselines. \textbf{MLP} is a straightforward collaborative filtering method: it produces node embeddings by using MLP and makes recommendations based on the dot product between embeddings. \textbf{NGCF}~\cite{wang2019neural} and \textbf{LightGCN}~\cite{he2020lightgcn} are state-of-the-art recommendation systems based on GCN~\cite{kipf2016semi}. Both methods propagate messages in a user-item graph to yield node embeddings; recommendations are made based on the dot product. The distinction between the two lies in their architectures: NGCF incorporates feature transformations and nonlinear activations inside its GCN layers, whereas LightGCN does not.

\subsubsection{Evaluation Metrics.}
To evaluate the performance of top-$k$ recommendation, we employ Hit Ratio (HR) and Normalized Discounted Cumulative Gain (NDCG)~\cite{he2015trirank}, which are widely used in collaborative filtering~\cite{he2017neural, wang2019neural}. HR counts the number of ground-truth links appearing in the top-$k$ ranked list and HDCG gives a higher score when the ground-truth links are ranked higher.

\subsubsection{Implementation Details.}
Both architectures of \revclassify can serve as the pretrained classifier for \revfilter; we use \revclassifyDS because of its faster inference and robustness (Section~\ref{sec:classification}). We also employ the two enhancements mentioned in Section~\ref{sec:discover}: we fine-tune the pretrained classifier with merging pairs, setting $\gamma = 0.4$, $1 \le n_{merge} \le 20$, and $\alpha_{keep} = 1.5$.

We implement NGCF and LightGCN using publicly available codebases, while incorporating the following changes for our setting. (1) We initialize the embedding layer with node features instead of using the Xavier initialization~\cite{glorot2010understanding}. (2) We optimize the binary cross entropy loss instead of the BPR loss~\cite{rendle2012bpr} due to the limited number of labeled links available.
We train the baselines for 150 epochs incorporating early stopping, with hyperparameter (number of layers, dropout, and hidden dimension) searching.

\subsection{Results}

\subsubsection{Performance Comparison With Baselines (Q1).}
Table~\ref{tab:recommendation} presents the comparison of methods in eight diverse settings of $\boldsymbol{n^++n^-@k}$. \revfilter outperforms all baselines on both metrics. In most cases, \revfilter achieves an HR that is 50--100\% higher than the second best method, indicating that we can identify 1.5 to 2 times more money-laundering schemes.

\subsubsection{Impact of Sparsity (Q2a).}
As shown in Figure~\ref{fig:rec_plot_sparsity}, \revfilter demonstrates significant robustness in sparse settings (i.e., low density), where there are few illicit links. For instance, when the density of $\tS$ decreases from $10^{-1}$ to $10^{-4}$ (i.e., 1000x more sparse), \revfilter experiences only a $\sim 20\%$ loss in HR, whereas NGCF and LightGCN show a $\sim 50\%$ drop and MLP suffers a $\sim 100\%$ decrease.

\subsubsection{Impact of $k$ (Q2b).}
We present the performance of the methods in the $1+1000@k$ setting, with varying $k$ in Figure~\ref{fig:rec_plot_k}. \revfilter can identify the ground-truth links in more than $90\%$ chance with a small number of recommendations ($k=10$), whereas the baselines cannot reach $75\%$ with a large number of recommendations ($k=100$). This result underscores that \revfilter can recommend a small yet high-quality set of suspicious subgraphs, significantly reducing the need for human analysts to examine numerous potential money laundering schemes. Note, unlike others, that our method's NDCG decreases as $k$ increases. This is because 1) as $k$ increases, our method's iterative filtering becomes less effective and 2) our method is optimized for filtering rather than ranking links. Therefore, it is important to select a proper $k$ to avoid suboptimal performance associated with an excessively large $k$.

\subsubsection{Iterative Versus One-Pass Filtering (Q3).}
To verify the usefulness of iterative filtering, we compare against a variant (``No iterations''). In this variant, the outer loop in Algorithm~\ref{alg:rec} is removed and the filtering is done in a single-pass, by computing scores for every 1-1 sender-receiver pair and recommending the top-$k$ pairs. The results in Table~\ref{tab:rec_ablations} show a significant drop in HR (0.4 to 0.6) for sparse $\tS$. This suggests that iterative filtering can mitigate the errors of the base classifier.

\subsubsection{Impact of Fine-Tuning With Data Augmentation (Q4a).}
We investigate the impact of fine-tuning the classifier with augmented data by comparing against the case without fine-tuning (``No fine-tuning''). Table~\ref{tab:rec_ablations} shows that the performance without fine-tuning drops across the board, with more significant decrease in sparse settings. This indicates that fine-tuning improves the classifier's accuracy, particularly for large sender-receiver pairs.

\subsubsection{Impact of Keeping More Candidates (Q4b.)}
To evaluate the effect of keeping more candidates on recommendation performance, we compare our method against the standard case of $\alpha_{keep} = 1$, where exactly $k$ candidates are retained in each iteration. The latter case leads to a slight decline in performance in most of the settings. We speculate that the fine-tuned classifier is already near-perfect, hence the improvement brought in by more candidates is minor. Since keeping more candidates compromises the inference speed, we suggest considering the trade-off when applying a large $\alpha_{keep}$.

\begin{table}[h]
  \centering
  \caption{Ablation (increasing sparsity from left to right).}
  \label{tab:rec_ablations}
  \begin{tabularx}{\columnwidth}{Xcccc}
    \toprule
    Method
    & 1+80@10
    & 1+640@10
    & 1+5120@10
    & 1+10240@10 \\
    \midrule
    \revfilter
    & $\boldsymbol{0.9710}$
    & $\underline{0.9128}$
    & $\boldsymbol{0.7969}$
    & $\boldsymbol{0.7161}$ \\
    No iterations
    & 0.9609
    & 0.5286
    & 0.1484
    & 0.1367 \\
    No fine-tuning
    & 0.8724
    & 0.6563
    & 0.3698
    & 0.3815 \\
    $\alpha_{keep} = 1$
    & $\underline{0.9688}$
    & $\boldsymbol{0.9167}$
    & $\underline{0.7891}$
    & $\underline{0.7122}$ \\
    \bottomrule
  \end{tabularx}
\end{table}

\section{Conclusions}
\ellipticII introduced a subgraph approach for AML and it set a new standard for forensic analysis in cryptocurrencies. While the original paper benchmarked a few subgraph classification methods and showed the promise of subgraph modeling, in this paper we advance the state-of-the-art by (1) improving the effectiveness of subgraph classification and (2) developing the capability of new subgraph discovery. \revclassify abstracts a transaction subgraph by using its initial senders and end receivers of the funds, significantly reducing the model inference cost and the memory consumption; while \revfilter discovers new subgraphs by using an innovative approach inspired by recommendation systems: recommending suspicious links between senders and receivers, which may not be directly connected by a single transaction. We have conducted many empirical evaluations to show that these two approaches significantly outperform strong baselines. An avenue of future research is to discover more complex money laundering schemes (e.g., those involving more than a single sender and receiver). Another avenue is to study the temporal behavior of money laundering and develop more effective methods that leverage both temporal and structural information.

\section{Supporting code}
Our implementation and pretrained models are available at \url{https://github.com/MITIBMxGraph/RevTrack}.

\begin{acks}
This work was funded by the MIT-IBM Watson AI Lab.
\end{acks}

\bibliographystyle{ACM-Reference-Format}
\bibliography{paper}

\end{document}